\documentclass[a4paper, 10pt, conference]{IEEEtran}      

\IEEEoverridecommandlockouts                              

\usepackage{graphics} 
\usepackage{graphicx}
\usepackage{caption}
\usepackage{subfigure}
\usepackage{float}
\usepackage{tabularx,booktabs}
\usepackage{amsmath} 
\usepackage{amssymb}  

\title{\LARGE \bf
Zero Initialization of modified Gated Recurrent Encoder-Decoder Network for Short Term Load Forecasting
}

\author{Vedanshu$^{1}$ and M M Tripathi$^{2}, ~\IEEEmembership{Senior member,~IEEE}$
\thanks{$^{1}$Vedanshu is with the Department of Electrical Engineering,
        Delhi Technological University, India
        {\tt\small vedanshu at outlook.com}}%
\thanks{$^{2}$M M Tripathi is with Faculty of Electrical Engineering,
        Delhi Technological University, India
        {\tt\small mmmtripathi at gmail.com}}%
}

\begin{document}

\maketitle
\thispagestyle{empty}
\pagestyle{empty}

\begin{abstract}

Single layer Feedforward Neural Network(FNN) is used many a time as a last layer in models such as seq2seq or could be a simple RNN network. The importance of such layer is to transform the output to our required dimensions. When it comes to weights and biases initialization, there is no such specific technique that could speed up the learning process. We could depend on deep network initialization techniques such as Xavier or He initialization. But such initialization fails to show much improvement in learning speed or accuracy. In this paper we propose  Zero Initialization (ZI) for weights of a single layer network. We first test this technique with on a simple RNN network and compare the results against Xavier, He and Identity initialization. As a final test we implement it on a seq2seq network. It was found that ZI considerably reduces the number of epochs used and improve the accuracy. The developed model has been applied for short-term load forecasting using the load data of Australian Energy Market. The model is able to forecast the day ahead load accurately with error of $0.94\%$.

\end{abstract}


\section{INTRODUCTION}

Recurrent Neural Network (RNN) are very powerful and is used for timeseries forecasting in various ways. But the problem with such type of network is that we always need a FNN as a last layer and no specific technique for parameter initialization exist for such layer. When we look at some of the works done in this field such as Xavier\cite{c1}, He\cite{c2} and identity matrix \cite{c3}, we see significant results with Rectified Linear Unit (ReLu) or its variants on a deep neural network. Other recent work include Layer-sequential unit-variance (LSUV)\cite{c4} initialization which also works great with ReLu family and deep network.  All such type of initialization were designed so that the input signal could traverse deep in the network. But using these techniques on a single layer network makes no sense. But we could still see these type of techniques being used on a single FNN. Thus, we propose a ZI for single FNN and tested it against various initialization techniques and implemented it on prize winning models. Results shows that ZI is way much superior than any other technique out there. We further investigate how a FNN does not die with both weights and biases initialized to zero and why it improves the accuracy. 

Electricity market has changed significantly in last decade forming a restructured market\cite{c17} where the prediction has become difficult due to emerging technologies. Short term load forecasting is particularly important for power system security and electricity cost. Mainly two types of forecasting is done, one is point forecasting and other is probability forecasting. If forecasting is done inaccurately it would lead to increased operating cost due to allocation of insufficient reserve capacity and use of expensive peak load units. Also, when talking about spot price and electrical energy trading in open market, electricity load forecasting plays  a major role to purchase the electricity at minimum cost. Point forecasting has emerged as the dominating technique for prediction of load with the major models being exponential smoothing\cite{c11} and regression\cite{c12,c17} in the field of statistical models.

When talking about artificial neural network\cite{c18}, models with variant of neural network have shown significant results in this field such as Improved Neural Network\cite{c10, c15}, Support Vector Machine \cite{c13} and Fuzzy System \cite{c14, c16}. There are two major problems associated with artificial neural network, namely, ``overfitting'' in which the model fails to train for the underlying relationship and the other is ``dimentionality'' in which the models complexity increases exponentially as the number of input dimension increases. 

The focus of this paper is to use a GRU encoder-decoder model to maximize the conditional probability of the target sequence given the input sequence. This model is mainly used in Neural Machine Translation where the input sequence length is variable in nature along with the output sequence length. In the decoder of the model, the fully connected dense neural network was initialized with the defamed Zero Initialization technique. We further investigate how initialization of both weights and biases to zero could be possible and why we didn't end up with dead neurons. It was shown that with only changing the initialization technique of this output layer we could get high level of accuracy in comparison to other initialization techniques. 

For testing of our models we have used electricity load of New South Wales (NSW). The timeseries is half hourly distributed and we have taken 48 hours as our forecasting horizon. We have two models and used various initialization techniques to compare our results. The results show that convergence speed and variance shown when the model initialized with ZI were superior to when initialized with other techniques. The mean absolute percentage error(MAPE) in load forecasting with ZI was around $0.94 \%$ while with Xavier Uniform was around $1.06\%$. Other techniques showed higher error and variance. 

In section II of this paper we discuss how the models were created and the pre-processing techniques used upon the dataset. In this section we discuss about the initialization technique itself and try to understand how ZI was able to give considerable results. Section III discusses the results obtained during testing of the model. And we conclude this paper with section IV. This paper also discusses in detail about the model and data pre-processing used to achieve high accuracy.


\section{Experiment setup}
\subsection{Dataset used}
For our models we have used electricity load of New South Wales (NSW)\cite{c5} of Australian Energy Market. The feature used from the dataset includes only total demands and its time. We have used historical electricity load data of year 2015,2016,2017. 

\subsection{Data pre-processing and feature engineering}
Raw input values was transformed by \emph{log(1 + x)}. We first try to find any type of auto-correlation within the data. The auto-correlation plot showed strong yearly and quarterly seasonality in data and we tried to use that to increase the accuracy. First thing we did with the data was to find the annual and quarter autocorrelation of the input dataset. But due to unevenness of input interval due to leap years, monthly length differences, autocorrelation was found considering neighbouring data also to reduce noise as follows:

\begin{equation*}
    corr = 0.5 * corr(lag) + 0.25 * corr(lag - 1) +  0.25 * corr(lag + 1)
\end{equation*}

 where, \emph{corr(lag)} is the autocorrelation with \emph{lag}. Two \emph{lag}s were used here; one of $365$ and other of $365.25/4$. These two values is then used as a feature. These were time independent features so it was necessary to stretch it to the timeseries length, i.e., they were added to at the end of each input batch. Next feature which we selected was day-of week. Since it could not be directly feed to the network, it was normalized as follows:

\begin{verbatim}
    feature = {mon: 0, ..., sun:6}
    
    for each feature_value in feature:
       normed = feature_value / (7 / 2*pi)
       dow = [cos(normed), sin(normed)]
\end{verbatim}

\emph{dow} from above was used as additional features. One of the important feature used was the \emph{lagged\_data}. Since data showed string quarter and annual correlation we decided to lag the input load by $3, 6, 9 ,12$ months and add them as additional features as follows:

\begin{verbatim}
    timeseries_indices: time indices of 
              the electricity load data
    offset: in months 
    
    for each time in timeseries_indices:
        for each offset in {3, 6, 9, 12}:
            date = time - offset
            lagged_data = data(date)
\end{verbatim}

\begin{figure}[H]
    \centering
    \includegraphics[width=\columnwidth]{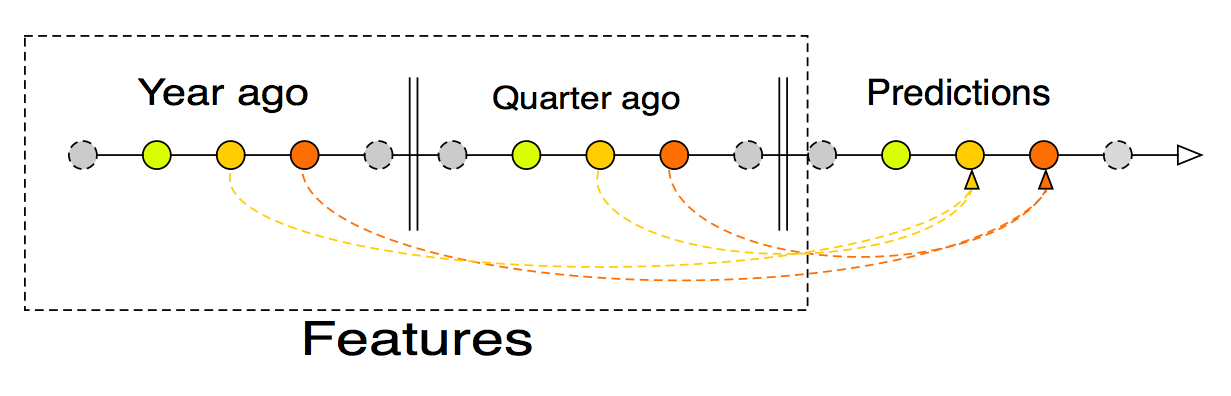}
    \caption{Quarter and annual feature selection for lagged\_data}
    \label{fig:from_past}
\end{figure}

Important features from the past are now explicitly included in the features and hence our model need not to remember very old informations. Lagged datapoints helps in reducing  the size of model and hence helps in faster training and less loss of information. This technique being simple helps in achieving high accuracy. Also, LSTM/GRU units tends to forget oldest informations when the input sequence size increases. 

\begin{figure}[H]%
\centering
\subfigure[][]{%
\label{fig:x_features}%
\includegraphics[width=0.45\textwidth]{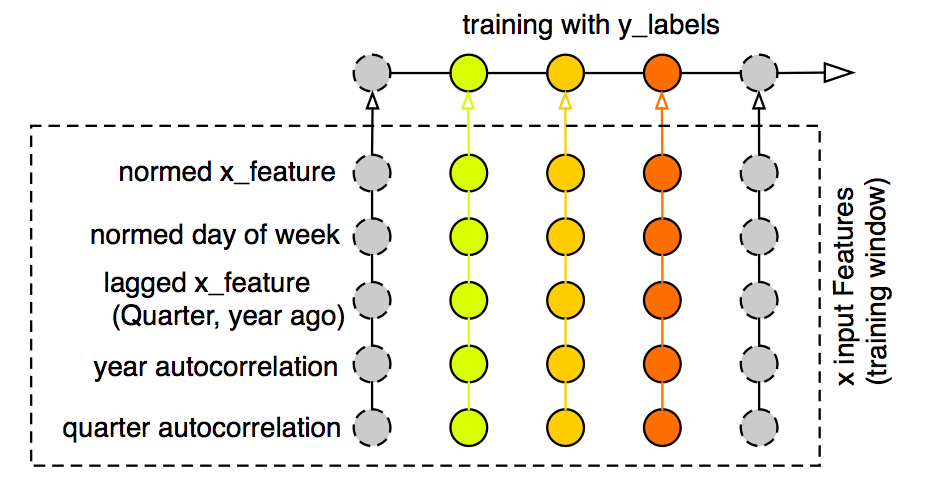}}%
\hspace{8pt}%
\subfigure[][]{%
\label{fig:y_features}%
\includegraphics[width=0.45\textwidth]{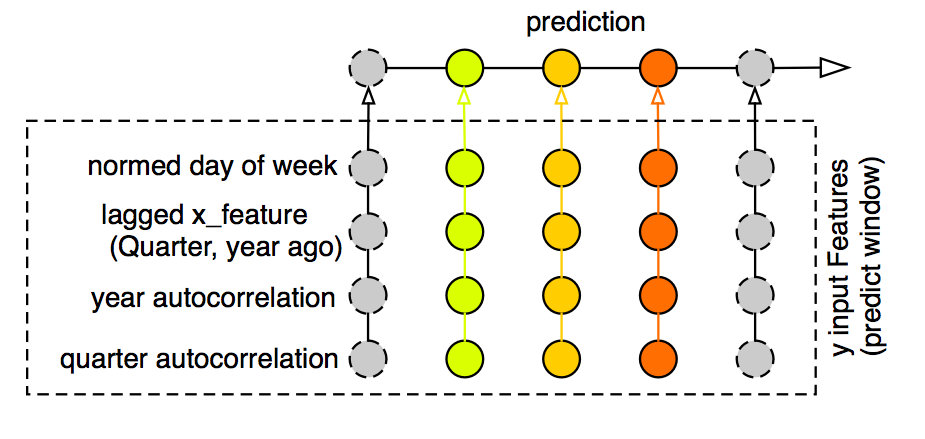}} %
\caption[A set of four subfigures.]{Visualization of input features:
\subref{fig:x_features} shows x input features used for training;
\subref{fig:y_features} shows y input features used for prediction.}
\label{fig:features}%
\end{figure}

Figure \ref{fig:features} shows how features were combined as input pipeline for being feed to the model. \emph{x\_features} were splitted into batches of size \emph{training\_window} and similarly \emph{y\_features} were splitted into batches of size \emph{predict\_window}. Year and quarter autocorrleation were tiled according to the size of input features. 


\subsection{Prediction Model 1}
\begin{figure}[H]
    \centering
    \includegraphics[width =\columnwidth]{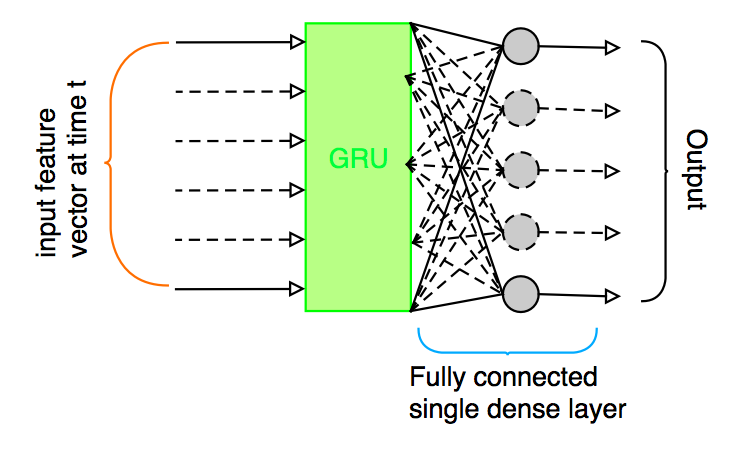}
    \caption{First model for electricity load forecasting with FC output layer}
    \label{fig:model1}
\end{figure}

We have done majority of the work to get the accuracy in the input pipeline so we tried to keep our first model as simple as possible. With this model we try to understand the concept of ZI and its significance. A simple RNN was selected and we used GRU \cite{c6} cells instead of LSTM. The output from the GRU layer is then connected to a fully connected (FC) dense layer. This is the layer where we will apply our initilization technique and compare the results with different initilization techniques. Figure \ref{fig:model1} shows the graphical depiction of our first model.

\subsubsection{Analysis of ZI}

Let us consider $w_{jk}^L$ to be the weight connecting $k^\text{th}$ neuron in the $(L-1)^\text{th}$ layer to the $j^\text{th}$ neuron in the $L^\text{th}$ layer. Here, $L^\text{th}$ layer is our last FC layer and  $(L-1)^\text{th}$ layer is the output layer of our GRU block. For simplicity, first let us consider identity function to be our activation function. The activation of the $j^\text{th}$ neuron of the last layer will be:

\begin{equation*}
    a_j^L = \sum_k (w_{jk}^{L-1} a_k^{L-1} + b_j^L)
\end{equation*}

We will consider quadratic cost function $C$ as follows:

\begin{equation}
    C  = \frac{1}{2n} \sum_x abs( y(x) - a^L(x) )^2
\end{equation}

where $n$ is the size of the training dataset. Also, $x$ is the input and $y = y(x)$ is the desired output. The weights are updated by finding tbe gradient of the cost function w.r.t. the weights as follows:

\begin{equation}
\label{w_update}
    w_{jk}^L \leftarrow w_{jk}^L - \eta \frac{\partial C}{\partial w_{jk}^L} 
\end{equation}

where, $\eta$ is the learining rate. From equation \ref{w_update} it can be seen that as long as the gradient is coming out to be non zero the weights would be updated and our neuron would not die. Since, we have used qudratic cost function it's derivative by the backpropogation equation \cite{c7} can be found easily as follows:

\begin{align}
    \delta_j^L &= \frac{\partial C}{ \partial a_j^L} \sigma'( z_j^L) \\
    \frac{\partial C}{ \partial a_j^L}  &= (a_j^L - y_j) \\
    \frac{\partial C}{\partial w_{jk}^L} &= a_k^{L-1} \delta_j^L
\end{align}

Since, our activation function was an identity, $\sigma'(z_j^L) = 1$. So, our updating gradient will be :

\begin{equation}
    \frac{\partial C}{\partial w_{jk}^L} = a_k^{L-1} * (a_j^L - y_j) * 1
\end{equation}

Since, our gradient came out to be non zero our neuron will keep on learning. Any activation function which have non zero derivative at zero will keep our neuron alive. Some of the alternative activation functions which could have been used are:

\begin{table}[H]
\begin{tabular}{rcl}
$a(z) = tanh(z)$    & ; & $a'(z) = sech^2(z)$ \\
$a(z) = sigmoid(z)$ & ; & $a'(z) = a(z)(1 - a(z))$            
\end{tabular}
\end{table}

\begin{figure}[H]%
\centering
\subfigure[][]{%
\label{fig:act1}%
\includegraphics[width=0.45\textwidth]{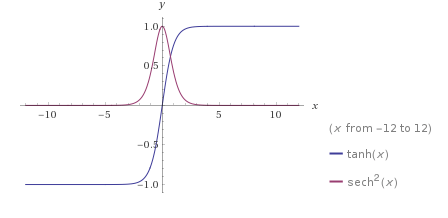}}%
\hspace{8pt}%
\subfigure[][]{%
\label{fig:act2}%
\includegraphics[width=0.45\textwidth]{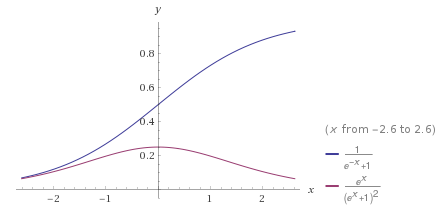}} %
\caption[A set of four subfigures.]{Plot of different activation function with their derivatives:
\subref{fig:x_features} $tanh(x)$ and $sech^2(x)$;
\subref{fig:y_features} $sigmoid(x)$ and $sigmoid(x)(1-sigmoid(x))$.}
\label{fig:act}%
\end{figure}

From figure \ref{fig:act} it's clear that the derivative of the activation functions are well defined at zero and are non-zero.

At the begining of the training all the weights and biases will be initialized to zero, so , $a_j^L = 0$. But $a_k^{L-1} \ne 0 $ since it is the output from the GRU block.So, it can be assumed that each neuron output coming from GRU will be different. With these assumptions the gradient at the begining will be 

\begin{equation}
    \frac{\partial C}{\partial w_{jk}^L} = a_k^{L-1} * (- y_j)
\end{equation}

Because of the different output coming from the GRU block, each neuron in FC layer will learn different weights with each iterations. Also, after training the weights are more tends towards zero, so it make sense to initialize all the weight to zero.


\subsection{Prediction Model 2}

\begin{figure*}[thp]
    \includegraphics[width=\textwidth]{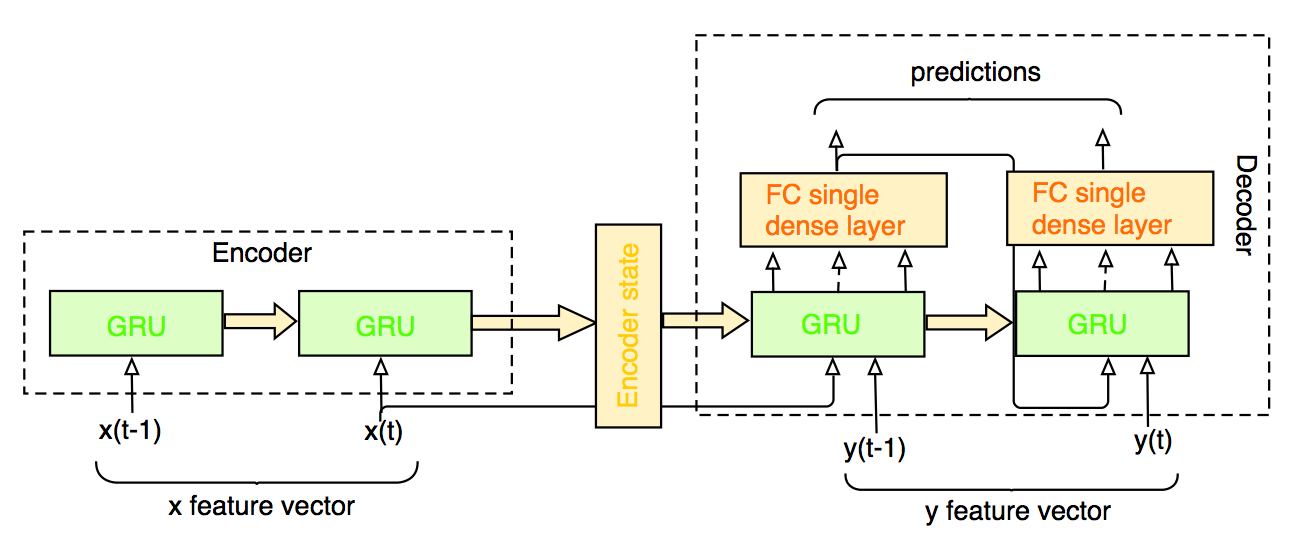}
    \caption{Second model for electricity load forecasting with FC output layer}
    \label{fig:model2}
\end{figure*}

A sequence to sequence (seq2seq) learning model \cite{c9} was selected to predict electricity load with high accuracy. This model was selected due to following reasons:
\begin{enumerate}
    \item A seq2seq model was used since our predictions is based upon conditional probability of previous values including our past predictions.
    \item This model is versatile with the type of features being injected into it, like numerical, categorical, timeseries etc.
\end{enumerate}
Figure \ref{fig:model2} depicts how the second model was constructed. 


\subsubsection{Losses and regularization}

\begin{figure}[H]%
\centering
\subfigure[][]{%
\label{fig:losses1}%
\includegraphics[width=0.45\textwidth]{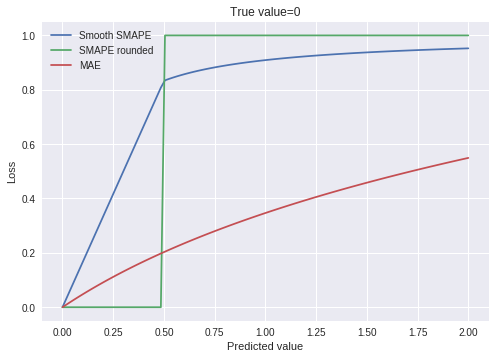}}%
\hspace{8pt}%
\subfigure[][]{%
\label{fig:losses2}%
\includegraphics[width=0.45\textwidth]{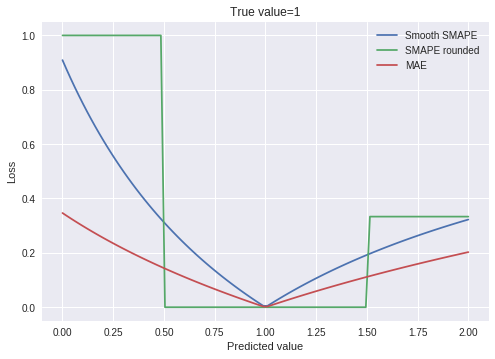}} %
\caption[A set of four subfigures.]{Plot of different loss functions:
\subref{fig:x_features} when True value = 0;
\subref{fig:y_features} when True value = 1.}
\label{fig:losses}%
\end{figure}

Smoothed Differential Symmetric Mean Absolute Percentage Error (SSMAPE) was used as a loss function in our model which is a variant of SMAPE. Mean Absolute Error (MAE) could have also been used on $log(1 + x)$. The reason for using these two was that their behaviour is well defined when truth value was close to zero and predicted value moves around zero. Figure \ref{fig:losses1} shows this problem graphically. The final predicted values were rounded to their nearest integer value and negative values were clipped to zero as can been seen in figure \ref{fig:losses}. 

\begin{equation}
    SMAPE = \frac{1}{N} \sum \frac{2*|F(t) - A(t)|}{|F(t)| + |A(t)|}
\end{equation}

\begin{equation}
    SSMAPE = \frac{1}{N} \sum \frac{2*|F(t) - A(t)|}{Max( |F(t)| + |A(t)| + \epsilon, 0.5 + \epsilon)}
\end{equation}

\begin{equation}
    MAE = \frac{1}{N} \sum |F(t) - A(t)|
\end{equation}

where $F(t)$ is the forecasted value and $A(t)$ is the actual value. $\epsilon$ is the smoothness factor in SSMAPE. 

Additional cost term for RNN were added to both encoder and decoder which adds squared magnitude of coefficient as penalty.

\begin{equation}
    L2 = \frac{\beta}{2} \sum R(t)^2 
\end{equation}
where, $R(t)$ is the RNN output and $\beta$ is the hyperparameter controlling the amount of regularization.

\subsection{Reducing model variance}

It was hard to know which training step would be best for predicting the future, so \emph{early stopping}\cite{c7} could not be used. So, it was necessary to use some technique to reduce the variance. Averaging Stochastic Gradient Descent (ASGD) was used to reduce variance in both model 1 and 2. With ASGD the moving average of the network weights are maintained during training. These averaged weights are used during predictions instead of the original ones.It has to be noted that if ASGD was used from the starting of training, the training would be very slow, so it was necessary to apply ASGD after certain epochs have been passed. ASGD was used only in FC layer for averaging of weights. 

\begin{figure}[H]
    \centering
    \includegraphics[width=\columnwidth]{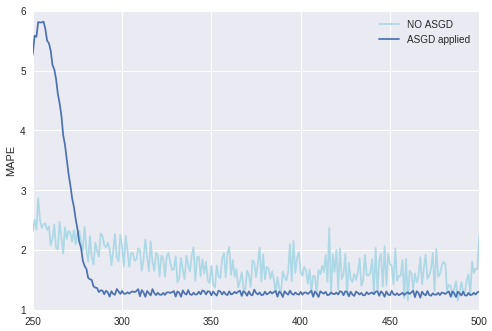}
    \caption{Variance reduction applied to the weights of FC layer of model 1}
    \label{fig:variance}
\end{figure}


\subsection{Training and validation}

The training and validation dataset was repeated within each epochs so as to reduce the number of epochs and improve the accuracy. The repetition of dataset was achieved as follows:

\begin{verbatim}
  data_train = data_train.repeat(n_repeat)
    
  for epoch in epochs:
    train(data_train)
\end{verbatim}

\section{Results and analysis}

\begin {table}
\caption{Model 1 results}
\label{tab:model1} 
\begin{tabularx}{\columnwidth}{lcc}
\toprule
    Initializer  & Average MAPE(\%)  & Standard deviation  \\ 
\midrule
Xa norm     & 9.0668       & 6.3522             \\
He          & 9.4730       & 6.7899             \\
Identity    & 2.9806       & 0.4611             \\
\textbf{ZI} & \textbf{1.7027} & \textbf{1.0745} \\
Xa uniform  & 10.258       & 8.0278             \\
\bottomrule
\end{tabularx}
\end {table}

\begin {table}
\caption{Model 2 results}
\label{tab:model2} 
\begin{tabularx}{\columnwidth}{lcc}
\toprule
    Initializer  & Average MAPE(\%)  & Standard deviation  \\ 
\midrule
Xa norm    & 1.6932 & 0.6014 \\
He         & 1.2288 & 0.4492 \\
Identity   & 1.6068 & 0.5823 \\
\textbf{ZI}& \textbf{0.9445} & \textbf{0.3918} \\
Xa uniform & 1.0579 & 0.3886 \\
\bottomrule
\end{tabularx}
\end {table}

Two variants of Xavier initialization were used in testing our models. The first one draws samples from normal distribution and the second one draws from uniform distribution within the $[-limit,limit]$. Where, $limit$ is defined for the normal distribution as,

\begin{equation}
    limit = \sqrt{\frac{2}{in + out}}
\end{equation}

and for uniform distribution as,

\begin{equation}
    limit = \sqrt{\frac{6}{in + out}}
\end{equation}

where, $in$ and $out$ are the number of input and output units in the weight vector \cite{c1}. Xavier uniform initializer is also called Glorot uniform initializer. 

The second initializer called He, draws samples from a truncated normal distribution within $[-limit, limit]$ where $limit$ is,

\begin{equation}
    limit = \sqrt{\frac{2}{in}}
\end{equation}

its variant draws samples from uniform distribution within $[-limit, limit]$

\begin{equation}
    limit = \sqrt{\frac{6}{in}}
\end{equation}

The Identity initializer used is a simple one which utilizes an identity matrix for initialization. 

All the initialization techniques where used for initialization of two models having single layer output network and the results are shown in table \ref{tab:model1} and \ref{tab:model2}. From table \ref{tab:model1} one can see that the best performance was given by the model when the model was initialized with zero. The results motivated us to test this technique on a complex network whose results are shown in table \ref{tab:model2}. From table \ref{tab:model2} the average MAPE we got with ZI was around $0.9445$ which is superior to any other MAPE given by other initialization techniques. The nearest to it was given by Xavier with uniform distribution, which was around $1.0579$. The difference between the errors with the two initialization techniques is not much but when compared to other initialization techniques the difference is quite comparable. It has to be noted that ZI is a simple technique of initialization and performs good on a single layer network when compared to other complicated initialization techniques. 

\begin{figure}[H]
    \centering
    \includegraphics[width=\columnwidth]{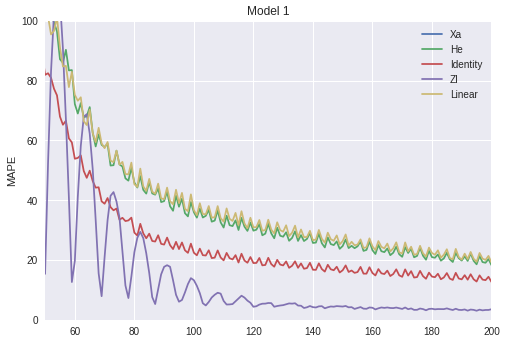}
    \caption{Training steps from model 1 using various initialization techniques}
    \label{fig:model1_results}
\end{figure}

To see how well the convergence is taking place during training we drew the plot during convergence in figure \ref{fig:model1}. Figure \ref{fig:model1} was drawn during training of our first model. From there one could see that the model is converging quite well with the ZI technique. 

\begin{figure}[H]
    \centering
    \includegraphics[width=\columnwidth]{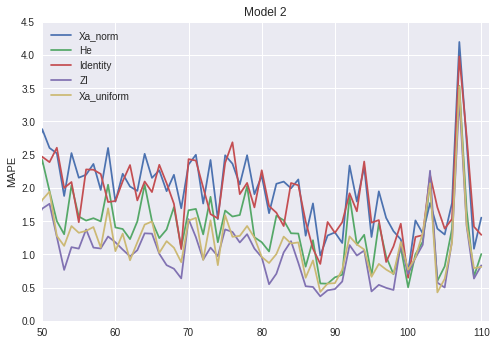}
    \caption{Plot of MAPE of each 48 hour of prediction of test set.}
    \label{fig:model2_mape}
\end{figure}

To dig a little dipper into testing we draw the plot of MAPE of each 48 hour of testing dataset which is shown in figure \ref{fig:model2_mape}. Figure \ref{fig:model2_mape} was drwan during testing of model 2. From it one can seen that the lowest MAPE was given by ZI technique for each 48 hour of testing therefore improving the overall MAPE of the model.

\section{Conclusion}

In this paper, we propose the use of zero initialization technique for single layer output of a complex network used for time series forecasting of electrcity load. The proposed model was able to reduce the variance and also increase the speed of training.

We evaluate our technique with other modern initialization techniques and model and it was shown that our technique was superior to other techniques. 
The model used was a seq2seq network since the output is based upon the conditional probability of previous output and input. Also, seq2seq network could be extended for other timeseries forecasting such as webpage views on a daily basis and stock price predictions. 

Before feeding the model with the input pipeline we did some preprocessing of data such as autocorrelation within the data. The dataset should strong annual and quarter correlations which was fed into the input pipelines. We also showed why some activation functions can't be used with ZI.



\begin{thebibliography}{99}

\bibitem{c1} Glorot X, Bengio Y. Understanding the difficulty of training deep feedforward neural networks. InProceedings of the thirteenth international conference on artificial intelligence and statistics 2010 Mar 31 (pp. 249-256).

\bibitem{c2} He K, Zhang X, Ren S, Sun J. Delving deep into rectifiers: Surpassing human-level performance on imagenet classification. InProceedings of the IEEE international conference on computer vision 2015 (pp. 1026-1034).

\bibitem{c3} Le QV, Jaitly N, Hinton GE. A simple way to initialize recurrent networks of rectified linear units. arXiv preprint arXiv:1504.00941. 2015 Apr 3.

\bibitem{c4} Mishkin D, Matas J. All you need is a good init. arXiv preprint arXiv:1511.06422. 2015 Nov 19.

\bibitem{c5} Australian Energy Market Operator, https://aemo.com.au/Electricity/National-Electricity-Market-NEM/Data-dashboard\#aggregated-data

\bibitem{c6} Cho K, Van Merriënboer B, Gulcehre C, Bahdanau D, Bougares F, Schwenk H, Bengio Y. Learning phrase representations using RNN encoder-decoder for statistical machine translation. arXiv preprint arXiv:1406.1078. 2014 Jun 3.

\bibitem{c7} Michael A. Nielsen, "Neural Networks and Deep Learning", Determination Press, 2015 


\bibitem{c9} Sutskever I, Vinyals O, Le QV. Sequence to sequence learning with neural networks. InAdvances in neural information processing systems 2014 (pp. 3104-3112).

\bibitem{c10} Rafiei M, Niknam T, Aghaei J, Shafie-khah M, Catalão JP. Probabilistic Load Forecasting using an Improved Wavelet Neural Network Trained by Generalized Extreme Learning Machine. IEEE Transactions on Smart Grid. 2018 Feb 21.

\bibitem{c11} Christiaanse WR. Short-term load forecasting using general exponential smoothing. IEEE Transactions on Power Apparatus and Systems. 1971 Mar(2):900-11.

\bibitem{c12} Ghelardoni L, Ghio A, Anguita D. Energy load forecasting using empirical mode decomposition and support vector regression. IEEE Trans. Smart Grid. 2013 Mar 1;4(1):549-56.

\bibitem{c13} Ceperic E, Ceperic V, Baric A. A strategy for short-term load forecasting by support vector regression machines. IEEE Transactions on Power Systems. 2013 Nov 1;28(4):4356-64.

\bibitem{c14} Sáez D, Ávila F, Olivares D, Cañizares C, Marín L. Fuzzy prediction interval models for forecasting renewable resources and loads in microgrids. IEEE Transactions on Smart Grid. 2015 Mar;6(2):548-56.

\bibitem{c15} Ertugrul ÖF. Forecasting electricity load by a novel recurrent extreme learning machines approach. International Journal of Electrical Power \& Energy Systems. 2016 Jun 1;78:429-35.

\bibitem{c16} Hassan S, Khosravi A, Jaafar J, Khanesar MA. A systematic design of interval type-2 fuzzy logic system using extreme learning machine for electricity load demand forecasting. International Journal of Electrical Power \& Energy Systems. 2016 Nov 1;82:1-0.

\bibitem{c17} Tripathi MM, Upadhyay KG, Singh SN. Short-term load forecasting using generalized regression and probabilistic neural networks in the electricity market. The Electricity Journal. 2008 Nov 1;21(9):24-34.

\bibitem{c18} Yadav HK, Pal Y, Tripathi MM. Photovoltaic power forecasting methods in smart power grid. InIndia Conference (INDICON), 2015 Annual IEEE 2015 Dec 17 (pp. 1-6). IEEE.

\end{thebibliography}
\end{document}